\documentclass{article}

\usepackage{microtype}
\usepackage{graphicx}
\usepackage{subfigure}
\usepackage{booktabs} 

\usepackage{hyperref}

\usepackage[accepted]{icml2023_deployableGenAI}

\usepackage{amsmath}
\usepackage{amssymb}
\usepackage{mathtools}
\usepackage{amsthm}

\usepackage[capitalize,noabbrev]{cleveref}

\theoremstyle{plain}

\theoremstyle{definition}

\theoremstyle{remark}

\usepackage[textsize=tiny]{todonotes}

\icmltitlerunning{Squeezing Large-Scale Diffusion Models for Mobile}

\begin{document}

\twocolumn[

\icmltitle{Squeezing Large-Scale Diffusion Models for Mobile}

\begin{icmlauthorlist}
\icmlauthor{Jiwoong Choi}{sqzb}
\icmlauthor{Minkyu Kim}{sqzb}
\icmlauthor{Daehyun Ahn}{sqzb}
\icmlauthor{Taesu Kim}{sqzb}
\icmlauthor{Yulhwa Kim}{snu}
\icmlauthor{Dongwon Jo}{snu}
\icmlauthor{Hyesung Jeon}{snu}
\icmlauthor{Jae-Joon Kim}{snu}
\icmlauthor{Hyungjun Kim}{sqzb}

\end{icmlauthorlist}

\icmlaffiliation{sqzb}{SqueezeBits Inc., Seoul, South Korea}
\icmlaffiliation{snu}{Seoul National University, Seoul, South Korea}

\icmlcorrespondingauthor{Hyungjun Kim}{hyungjun.kim@squeezebits.com}

\icmlkeywords{Machine Learning, ICML}

\vskip 0.3in
]

\printAffiliationsAndNotice{}  

\begin{abstract}
The emergence of diffusion models has greatly broadened the scope of high-fidelity image synthesis, resulting in notable advancements in both practical implementation and academic research.
With the active adoption of the model in various real-world applications, the need for on-device deployment has grown considerably. 
However, deploying large diffusion models such as Stable Diffusion with more than one billion parameters to mobile devices poses distinctive challenges due to the limited computational and memory resources, which may vary according to the device. 
In this paper, we present the challenges and solutions for deploying Stable Diffusion on mobile devices with TensorFlow Lite framework, which supports both iOS and Android devices. 
The resulting Mobile Stable Diffusion achieves the inference latency of smaller than 7 seconds for a 512 $\times$ 512 image generation on Android devices with mobile GPUs.
\end{abstract}

\section{Introduction}
\label{introduction}
Recently, diffusion models have gained significant interest by achieving impressive performance in image synthesis and related tasks.
Since the public release of Stable Diffusion~\citep{rombach2022ldm}, one of the foundation models in diffusion models, there has been a surge of interest in exploring the potential of the diffusion models in various fields including image synthesis \citep{ho2020ddpm,song2021ddim,rombach2022ldm,ho2022classifier,saharia2022photorealistic}, super-resolution \citep{li2022srdiff,sahak2023denoising,gao2023implicit}, inpainting \citep{lugmayr2022repaint,nichol2022glide,avrahami2022blended,gao2023implicit}, and many other applications \citep{luo2023videofusion,blattmann2023align,yang2023diffsound,liu2023audioldm}. 

Deploying large diffusion models on mobile devices offers significant advantages such as reduced server costs and improved user privacy, but it presents unique challenges.
These challenges arise from the large number of parameters, typically exceeding one billion, which necessitates compressing the model for deployment on mobile devices.
Moverover, ensuring that the computation latency remains within an acceptable range is also a crucial consideration.

In this paper, we introduce the implementation of Mobile Stable Diffusion based on the Stable Diffusion v2.1, achieving the lowest inference latency on GPU-powered Android devices, to the best of our knowledge ($\sim$7 seconds on Samsung Galaxy S23 to generate a 512 $\times$ 512 image).
\section{Background}
\label{background}
Diffusion models utilize the reverse diffusion process to generate images from noise. 
These models have been recognized for their ability to address significant challenges in the field of image synthesis.
Specifically, they mitigate problems such as mode-collapse, training instability, and quality degradation that are commonly encountered in previous approaches such as Generative Adversarial Networks (GANs) or Variational Autoencoders (VAEs).
\citet{ho2020ddpm} initially showcased the capability of diffusion models in generating high-quality images, although they came with high computational costs.
Subsequent works \citep{song2021ddim,rombach2022ldm} have focused on reducing the computational cost of diffusion models. 
\citet{song2021ddim} introduced a method to decrease the number of denoising steps based on the non-Markovian diffusion process.
On the other hand, \citet{rombach2022ldm} proposed to improve efficiency of diffusion models by applying denoising steps on latent space.

\begin{figure*}[ht]
\begin{center}
\centerline{\includegraphics[width=0.95\textwidth]{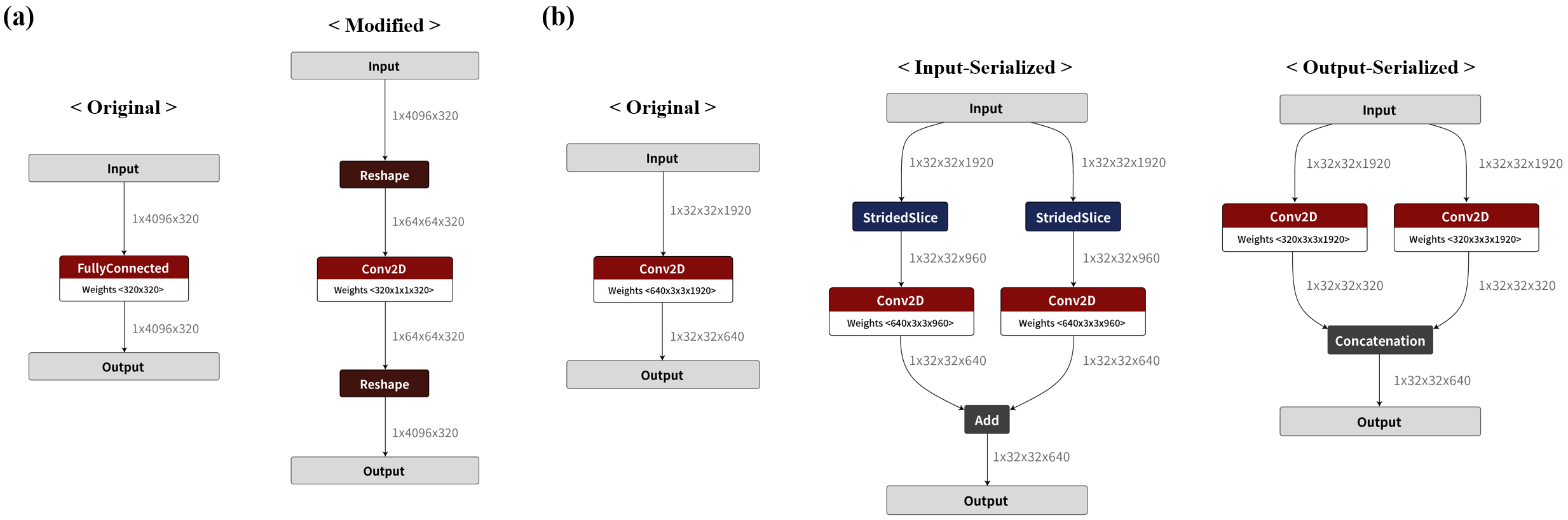}}
\vskip -0.12in
\caption{(a) Converting a fully-connected layer into a Conv2D layer. (b) Input- and Output-Serialization of a large Conv2D layer.}
\label{fig:conversion-and-serialization}
\end{center}
\vskip -0.32in
\end{figure*}

The advancement in improving efficiency in diffusion models contributed to the development of Stable Diffusion, a latent diffusion model for high-resolution image generation.
Stable Diffusion has demonstrated impressive capabilities in both text-to-image and image-to-image synthesis tasks. 
The model combines three modules to implement text-to-image synthesis; a Contrastive Language–Image Pre-training (CLIP) module that generates guidance from a given text prompt (text encoder), a U-Net module that conducts the reverse diffusion process (denoising network), and a Decoder module from a VAE model that generates an image from the output latent tensor (image decoder).

There is a growing demand for on-device image synthesis using the diffusion models, with a focus on enhancing the models in terms of latency, scalability, and user privacy.
\citet{apple} introduced the official support for on-device computations of Stable Diffusion on iOS mobile devices.
On Android devices, \citet{qualcomm} recently announced the first mobile deployment of Stable Diffusion based on the Hexagon processor of the latest Snapdragon 8 Gen 2 platform.
\citet{chen2023speed} has also demonstrated a faster implementation of Stable Diffusion using mobile GPUs based on private OpenCL kernels.
While prior works have demonstrated the feasibility of deploying Stable Diffusion on-device, these works commonly relied on custom-built kernels for acceleration.
Particularly in the case of Android devices, \citet{qualcomm} relied on the Hexagon processor and the dedicated SDK.
Additionally, \citet{chen2023speed} reported extensive use of private OpenCL-based kernels, pursuing additional performance gain with optimized memory access and faster computation.

\section{Challenges and Proposed Solutions}

We have chosen Google's TensorFlow Lite (TFLite) runtime \citep{tensorflowlite} as our deployment framework, rather than constructing custom-built kernels.
Opting for TFLite offers two significant benefits over building custom kernels.
First, the publicly accessibility of TFLite is likely to stimulate further adoption of on-device Stable Diffusion models in real-world applications.
Moreover, the versatility of TFLite facilitates the rapid deployment of various diffusion models on different mobile devices using the same optimization techniques.
In this section, we introduce several technical challenges we encountered while deploying the Stable Diffusion model using TFLite on a mobile GPU and propose solutions for them.

\subsection{Complete Mobile GPU Delegation}
TFLite enables the use of the mobile GPU via a hardware driver called GPU delegate.
It selectively runs supported operators in a computation graph on the GPU, leaving the unsupported operators to run on the CPU.
However, such selective execution often leads to sub-optimal performance due to the expensive communication between CPU and GPU.
Therefore, complete delegation is necessary for achieving optimal performance.
 
While the TFLite GPU delegate provides the acceleration for the most operators involved in Stable Diffusion, it fails to delegate even officially supported operators when the input activation size is large.
To address the incomplete GPU delegation, we propose three methods involving modifications in the computation graph of the model.

\subsection*{Converting $FullyConnected$ to $Conv2D$}
In spatial transformer blocks of the denoising U-Net network, there exist several fully-connected layers with large input activations (e.g., $1 \times 4096 \times 320$).
Since the large fully-connected layers failed to be delegated, we convert them to equivalent convolution layers as shown in Fig.~\ref{fig:conversion-and-serialization}.
Note that the depicted $FullyConnected$ layer and the $Reshape$-$Conv2D$-$Reshape$ layers result the same output and show almost the same latency when benchmarked on the GPU.
Hence, converting all $FullyConnected$ operators into equivalent $Conv2D$ operators is preferable to prevent the GPU delegation failure.

\subsection*{Serializing $Conv2D$ with large activations}

\begin{figure}[ht]
\begin{center}
\centerline{\includegraphics[width=0.9\columnwidth]{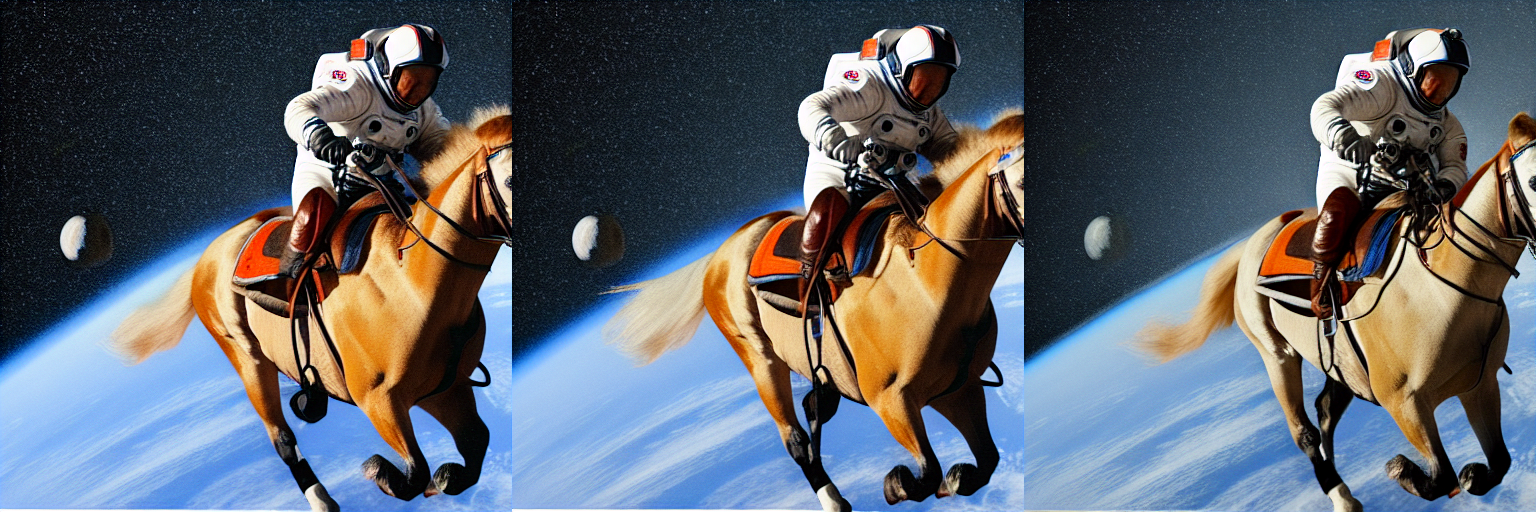}}
\vskip -0.12in
\caption{Images generated with the same textual description and initial latent with 20 iterations. From left to right: baseline, after applying input serialization for Conv2D, numerically stable GELU approximation on Macbook M1 Pro.}
\label{fig:compare-serialization}
\end{center}
\vskip -0.35in
\end{figure}

Although converting fully-connected layers to equivalent convolution layers enables delegation of layers with large input activations, we observed that one $3\times3$ convolution layer in the denoising network failed to be delegated with OpenCL backend due to its large input and output activation sizes: $1 \times 32 \times 32 \times 1920$ and $1 \times 32 \times 32 \times 640$, respectively.

Serializing the $Conv2D$ operator can solve this problem by reducing the activation sizes, but at the cost of multiple kernel call overhead. Therefore, the minimal serialization factor should be chosen to avoid excessive overhead.

The serialization can be applied along the input or output channel dimension as shown in Fig.~\ref{fig:conversion-and-serialization}.
We find that the minimal serialization factor that enables complete delegation is 2 with the latency of 15.5 ms for the input dimension, and 8 with the latency of 40.9 ms for the output dimension by trying possible serialization factors in increasing order along each dimension.
Thus, we chose the input serialization for its lower latency.

As the input serialization is a simple reordering of the computation sequence, the output should be very similar to that of the original graph. We qualitatively examined the generated images before and after applying the serialization.
The difference between the images was subtle, as shown in the first two images in Fig.~\ref{fig:compare-serialization}.

\subsection*{Broadcast-free Group Normalization}
Group normalization is not represented as a single operator in the TFLite but as a computation graph consisting of basic operators such as $Mean$, $Square$, $Rsqrt$, and $BroadcastTo$.
However, $BroadcastTo$ is not supported by the TFLite GPU delegate, which makes it necessary to modify the implementation of the group normalization layer.

We notice that the TFLite converter does not create an explicit $BroadcastTo$ operator when the activations are 4-dimensional or lower tensors. 
Hence, we reformat the group normalization layer so that the dimensions of the activation tensors are at most 4. 
Please refer to Fig.~\ref{fig:groupnorm} in Appendix for the modified group normalization graph.

\subsection{Numerically Stable Approximation of GELU}
\begin{figure}[ht]
\begin{center}
\centerline{\includegraphics[width=0.7\columnwidth]{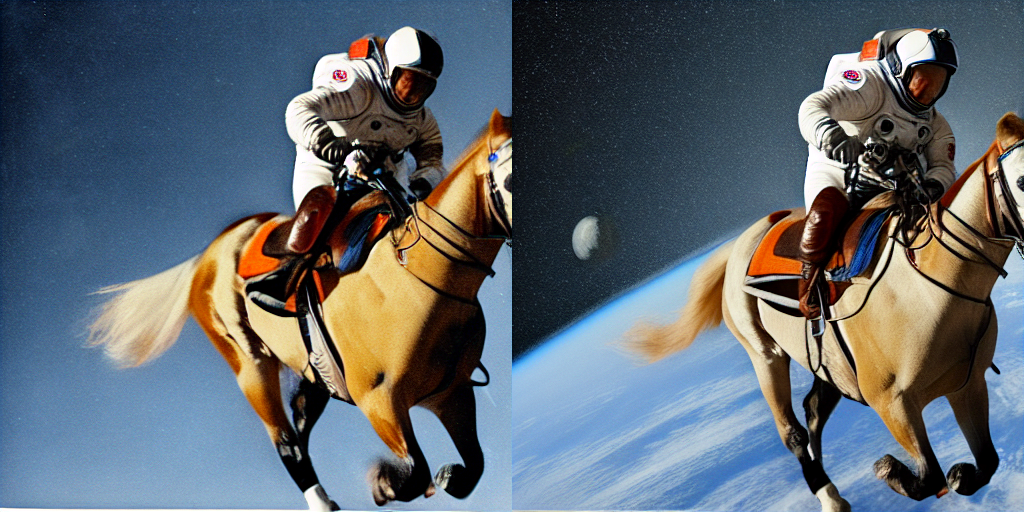}}
\vskip -0.12in
\caption{The images generated by different hardwares with the same initial latent and textual description with 20 iterations. [left: Galaxy S23 Ultra, right: Apple M1 Pro] }
\label{fig:mac-vs-android}
\end{center}
\vskip -0.3in
\end{figure}

The images generated on different hardwares are noticeably different even if identical textual description and initial latent have been used as inputs (Fig.~\ref{fig:mac-vs-android}).

Stable Diffusion adopts float16 as the default data type for faster operations, which generally works well on server GPUs without causing any issues. However, it is important to note that on certain mobile devices, the use of float16 can lead to floating-point exceptions.
We identify that the numerical instability is caused by the approximated $GELU$ operator in its cubic polynomial term.
$$GELU(x) \approx 0.5x(1 + \tau(x))$$
where $\tau(x) \coloneqq tanh\bigg(\sqrt{\frac{2}{\pi}}(x + 0.044715 x^3)\bigg)$

Instead of this well-known approximation, we use the following more numerically stable approximation:
$$GELU(x) \approx 0.5x\big(1 + \tau\big(\gamma_M(x)\big)\big)$$
where
$$\gamma_M(x) \coloneqq \begin{cases} x,\ \text{if}\ \  \lvert x \rvert \le M & \\ M,\ \text{otherwise} \end{cases}$$
is a clipping function.
We use an empirical value $M=10$, which suppresses the floating-point exceptions and maintains the image quality as shown in Fig.~\ref{fig:compare-serialization}.

\begin{table*}[t]
\caption{Comparison with different Stable Diffusion on Mobile. (image resolution: $512 \times 512$)}
\label{table:comparison}
\begin{center}
\begin{small}
\begin{sc}
\begin{tabular}{cccccc}
\toprule
& Model & Latency & Device & Hardware & Engine\\
\midrule
\citet{qualcomm} & SD v1.5 & $\sim$ 15s & (Galaxy S23) & Hexagon Proc. & Qualcomm AI Engine\\
\citet{chen2023speed} & SD v1.4 & $\sim$ 12s & Galaxy S23 Ultra & Mobile GPU & Custom Kernel\\
Ours & SD v2.1 & $\sim$ 7s & Galaxy S23 & Mobile GPU & TFLite\\
\bottomrule
\end{tabular}
\end{sc}
\end{small}
\end{center}
\vskip -0.1in
\end{table*}

\subsection{Pipelined Execution}
\begin{figure}[ht]
\begin{center}
\centerline{\includegraphics[width=\columnwidth]{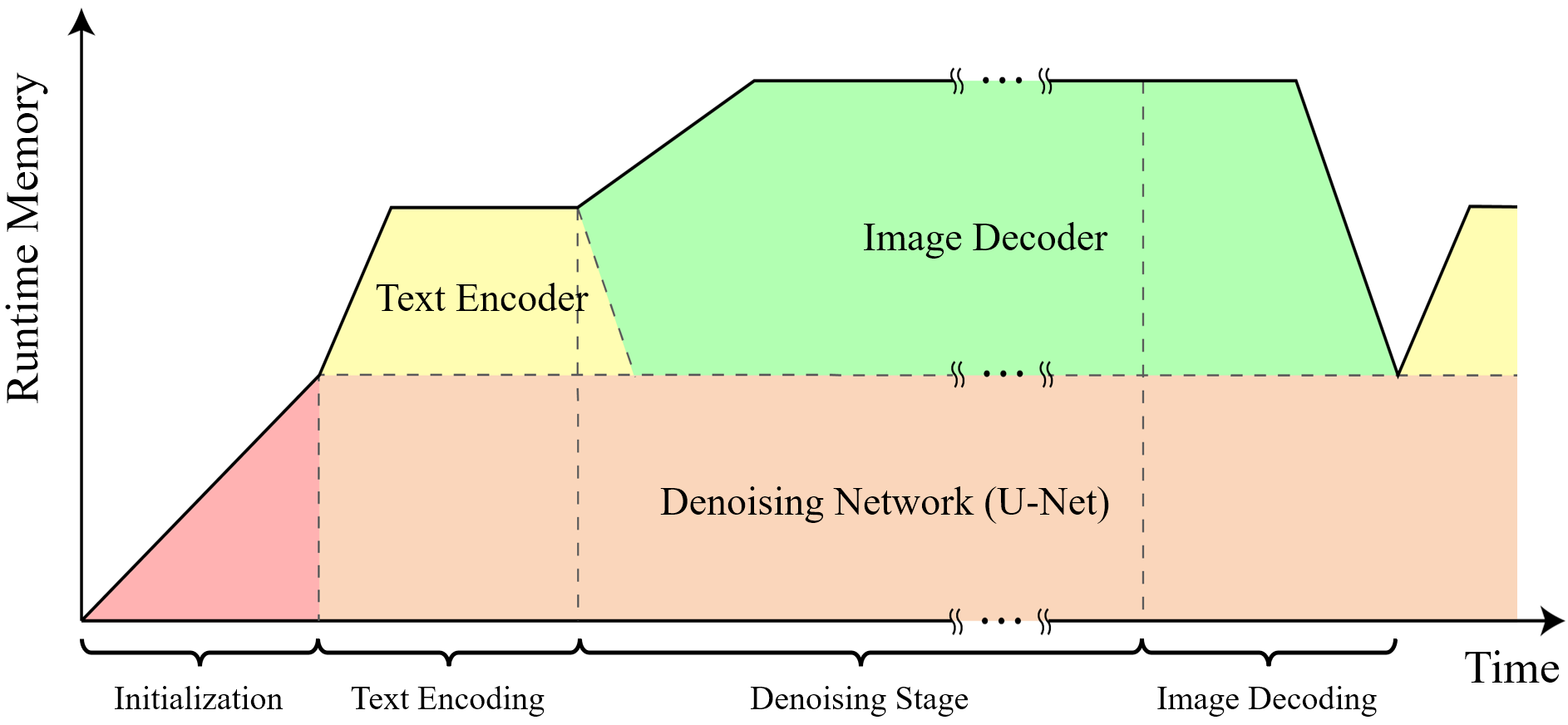}}
\vskip -0.12in
\caption{A qualitative illustration of the memory occupation of each component of Stable Diffusion during the pipelined execution. The orange (resp. yellow, green) area represents the memory occupied by the denoising network (resp. text encoder, image decoder).}
\label{fig:pipelined-execution}
\end{center}
\vskip -0.2in
\end{figure}
Due to the limited memory available on the mobile devices, it is often not practical to load all three components of Stable Diffusion on the memory simultaneously.

We propose a pipelined execution strategy for devices with small processor memory.
While the denoising network is retained on the memory throughout the entire execution, the text encoder and the image decoder are loaded interchangeably via a child thread running parallel with the main thread, as described in Fig.~\ref{fig:pipelined-execution}.

\subsection{Model Compression}
We apply quantization and pruning techniques to the pre-trained model to reduce the overall memory consumption.
Since mobile GPU does not support integer matrix multiplications, float16 is applied for the activations.
However, we quantize weights into 8-bit precision to reduce the model size; thus, weights are casted from 8-bit integers to 16-bit floating points before being involved in the computation.
We further apply structured pruning on huge convolution layers to minimize memory requirements.

\begin{figure}[ht]
\begin{center}
\centerline{\includegraphics[width=0.9\columnwidth]{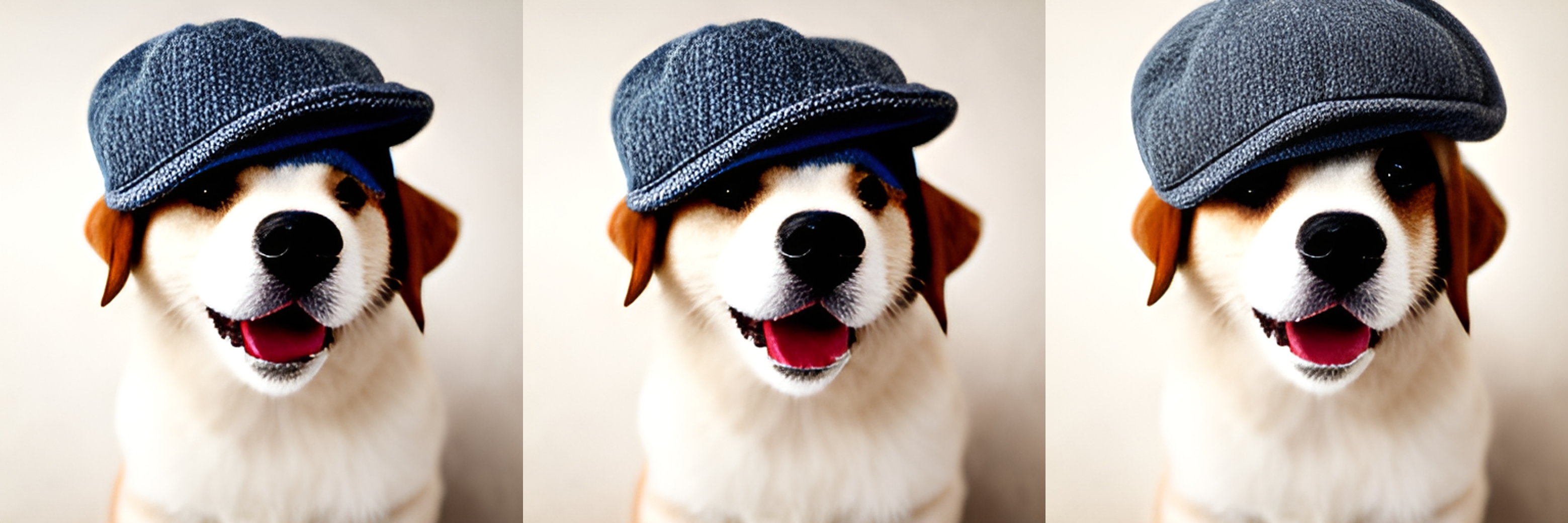}}
\caption{From left to right: baseline, after applying 8-bit weight quantization, and pruning.}
\label{fig:quantized-and-pruned}
\end{center}
\vskip -0.2in
\end{figure}

Since it is not straightforward to measure the performance degradation caused by the quantization and pruning quantitatively, we used block-wise reconstruction error~\cite{li2021brecq,wei2022qdrop} as an indirect metric and the quality of generated images as a qualitative measure.
Fig.~\ref{fig:quantized-and-pruned} shows the output images of the baseline, quantized, and quantized and pruned model, respectively.
Although each image shows differences in details, they are less prominent than in Fig.~\ref{fig:mac-vs-android}.

\section{Experiment}

In this work, we use Stable Diffusion v2.1 as a baseline model and optimize it for mobile deployment.
We choose Samsung Galaxy S23 device to measure end-to-end benchmark latency.
The device has Snapdragon 8 Gen 2 processor which includes Adreno 740 GPU.
In addition to the quantization and pruning, we apply knowledge distillation to reduce the number of inference steps following \citet{salimans2022distill} and \citet{meng2023distillation}.

\begin{figure}[h]
\begin{center}
\centerline{\includegraphics[width=0.95\columnwidth]{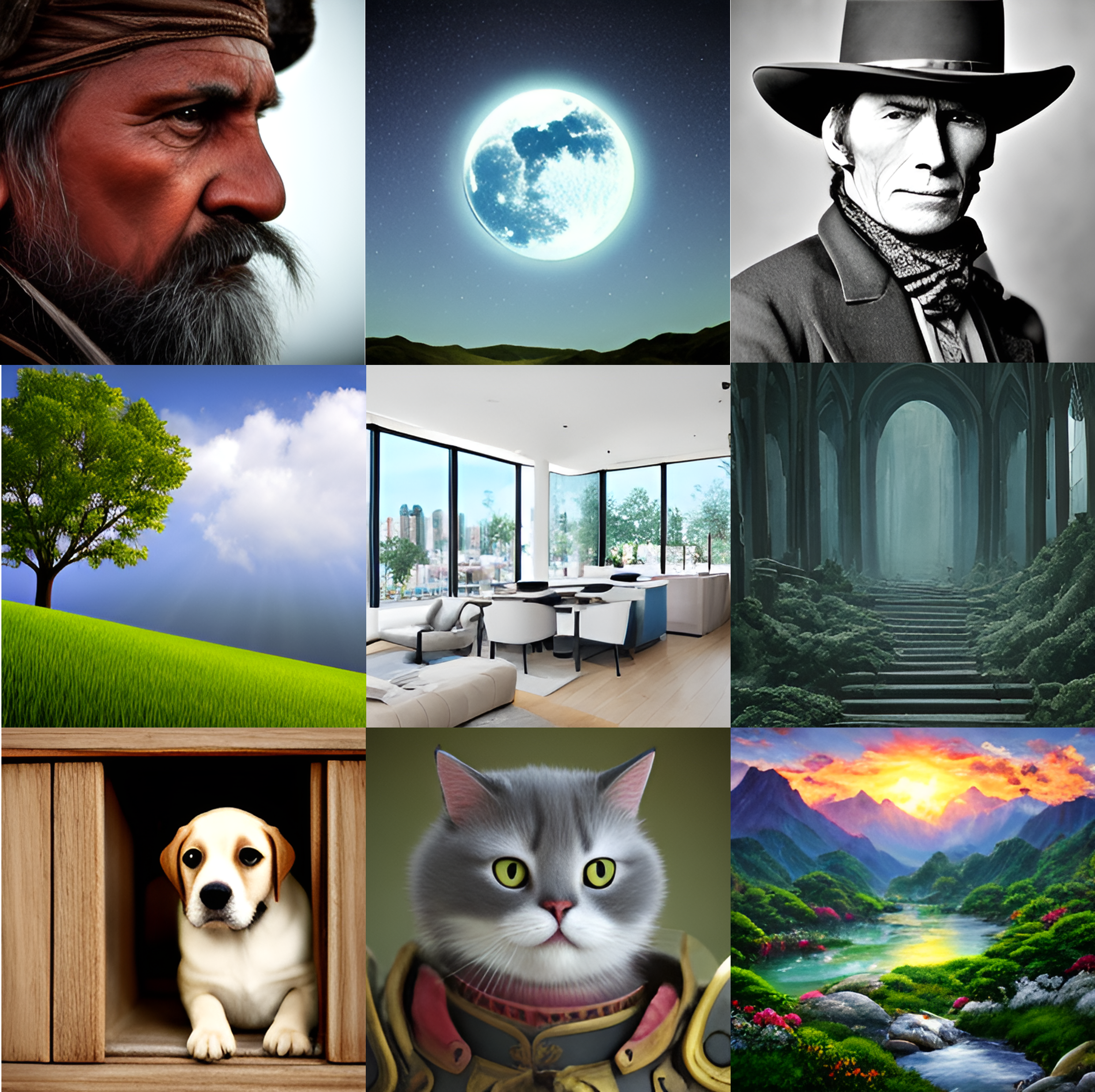}}
\caption{Example images generated by our method on a mobile device.}
\label{fig:example-images}
\end{center}
\vskip -0.2in
\end{figure}

Table~\ref{table:comparison} shows the end-to-end latency of our model and the comparison with previous approaches to deploy Stable Diffusion on mobile.
For a fair comparison with previous works, we measure end-to-end latency for text encoding, 20 effective denoising steps and image decoding.
The proposed approach can successfully generate a 512x512 image from a given text prompt within 7 seconds as shown in Fig.~\ref{fig:example-images}.
In addition, while previous approaches use dedicated or custom engine to deploy Stable Diffusion on mobile, our approach enables using off-the-shelf TFLite engine without any custom modification.

\section{Conclusion}
In this paper, we have discussed a series of optimization techniques that, in combination, enable the fastest on-device image synthesis using the Stable Diffusion.
These solutions can be extended to the deployment of other diffusion models, thereby facilitating the implementation of these models on various mobile devices, while leveraging the computation capability of TFLite.
We believe that the optimized deployment to a common and accessible inference framework will enrich the ecosystem of real-world mobile applications built upon diffusion models.

\bibliography{bibliography}
\bibliographystyle{icml2023}

\newpage
\appendix
\onecolumn
\section{Visualization of computational graphs}

\begin{figure*}[ht]
\begin{center}
\centerline{\includegraphics[width=0.73\textwidth]{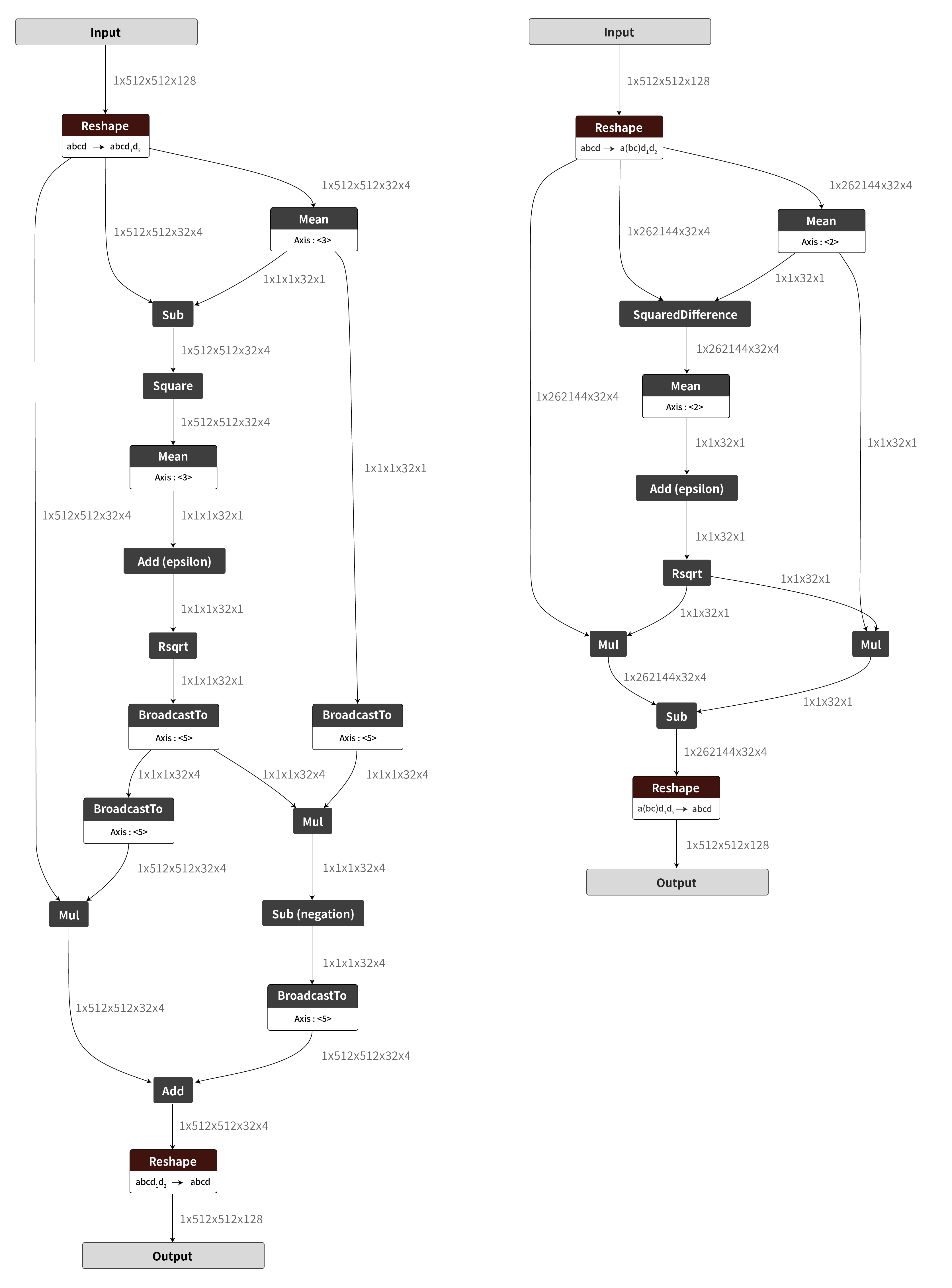}}
\caption{Left: the original group normalization; Right: reimplemented group normalization without any $BroadcastTo$ operator}
\label{fig:groupnorm}
\end{center}
\vskip -0.2in
\end{figure*}

We provide visualization of computational graphs proposed in the main text.
Fig.~\ref{fig:groupnorm} shows the computational graph of original group normalization layer in TFLite format and that of the reimplemented group normalization layer.
All of the $BroadcastTo$ operations and 5-dimension activations are removed in the reimplemented version.

\begin{figure}[ht]
\begin{center}
\centerline{\includegraphics[width=0.45\textwidth]{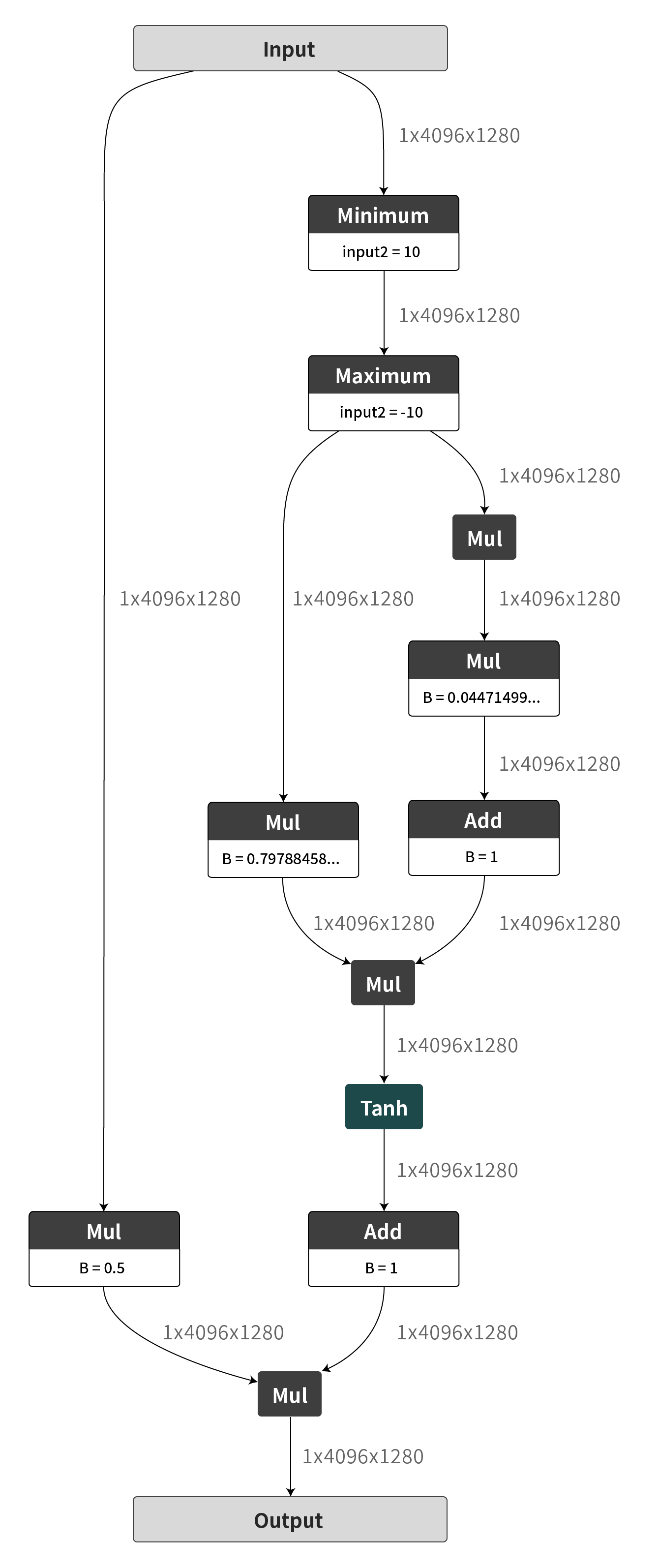}}
\caption{The numerically stable approximation of GELU}
\label{fig:safe-gelu}
\end{center}
\vskip -0.2in
\end{figure}

In Fig.~\ref{fig:safe-gelu} , the computational graph of the modified version of GELU is depicted.
Note that the additional operations ($Minimum$ and $Maximum$) are added in the beginning of the graph.

\end{document}